\documentclass[11pt]{article}

\usepackage{a4wide}
\usepackage{latexsym}
\usepackage{graphicx}
\usepackage{color}
\usepackage[hypertex]{hyperref}

\title{A Methodology for Learning Players' Styles from Game Records}

\author{Mark Levene and Trevor Fenner\\
School of Computer Science and Information Systems \\
Birkbeck College, University of London \\
London WC1E 7HX, U.K. \\ \{mark,trevor\}@dcs.bbk.ac.uk}

\date{}

\begin{document}

\maketitle

\begin{abstract}

We describe a preliminary investigation into learning a Chess player's style from game records. The method is based on attempting to learn features of a player's individual evaluation function using the method of temporal differences, with the aid of a conventional Chess engine architecture.
Some encouraging results were obtained in learning the styles of two recent Chess world champions, and we report on our attempt to use the learnt styles to discriminate between the players from game records by trying to detect who was playing white and who was playing black.
We also discuss some limitations of our approach and propose possible directions for future research. The method we have presented may also be applicable to other strategic games, and may even be generalisable to other domains where sequences of agents' actions are recorded.

\end{abstract}


\section{Introduction}

In Chess, as in other popular strategic board games, players have different styles. For example,
in Chess some players are more ``positional'' and other more ``tactical'', and this difference
in style will affect their move choice in any given board position, and more generally their overall plan. The problem we tackle in this paper is that of applying machine learning to teach a computer to discriminate between players based on their style. Before we explain our methodology, we briefly review the method of temporal difference learning, which is central to our approach.

\medskip

Temporal difference learning \cite{SUTT88} is a machine learning technique, originating from the seminal work of Samuel \cite{SAMU59}, in which learning occurs by minimising the differences between predictions and actual outcomes of a temporal sequence of observations. Samuel \cite{SAMU59} used the game of Checkers as a vehicle to study the feasibility of a computer learning from experience. Although the program written by Samuel did not achieve master strength, it was the precursor of the Checkers program Chinook \cite{SCHA97,SCHA01}, which was the first computer program to win a match against a human world champion. (See \cite{HOLL98} for a detailed, but less technical, description of the machine learning in Samuel's Checkers program.)  Tesauro \cite{TESA92} demonstrated the power of this technique by showing that temporal difference learning, combined with using a neural network, can enable a program to learn to play Backgammon at an expert level through self-play. Following this approach, there have been similar efforts in applying this technique to the games of Chess \cite{THRU94,BAXT00,BEAL00,BJOR03,MANN04}, Go \cite{SCHR94,SILV07}, Othello \cite{LEOU95,BINK07} and Chinese Chess \cite{TRIN98}. Self-play is time consuming, so it is natural to try to make use of existing game records of strong players to train the evaluation function, as in \cite{MANN04} (in which, however, the temporal difference training did not employ minimax lookahead). Learning from game records has also been used in the game of Go \cite{KOJI01,VAND05,STER06} to extract patterns for move prediction, using methods other than temporal difference learning.

\medskip

Here our aim is not necessarily to train a computer to be a competent game player, but rather to teach it to play in the style of a particular player, learning this from records of games played by that player. (In principle, the system could learn by interacting with the player but, when sufficient game records exist, learning can generally be accomplished faster and more conveniently off-line.) It is important to note that information available during learning should {\em not} include any meta-features such as the date when the game was played, the name of the opening variation played, or the result of the game. All the learning module observes is the sequence of moves played in each game.

\smallskip

Looking at it from a different perspective, we can view the problem as one of classification. Assume that we train the computer to play in the styles of two Chess players, say Kasparov and Kramnik. The problem can then be reformulated as follows: by inspecting the record of a game played between Kasparov and Kramnik, can the computer detect, with some confidence, which player was playing with the white pieces and which with the black pieces?

At an even higher level, the problem can be recast as a Turing test for Chess \cite{PELL08}, where a computer may fool a human that it is a human player. In some sense this may already be true for the strongest available computer Chess programs \cite{KROL99,ROSS03}, as computers have already surpassed humans in their playing strength, mainly due to increased computing power and relying on brute-force calculations.
Moreover, there seems to be a high correlation between the choices made by top human chess Grandmasters and world class chess engines (see \cite{CHES06a}).

\smallskip

We will not discuss the Turing test debate further and, from now on, we will concentrate on the classification problem within the domain of Chess. As far as we know, this is a new problem, and in this paper we suggest tackling it using temporal difference learning.  All previous uses of temporal difference learning in games (some of which are cited above) attempt to learn the weights of an evaluation function in order to improve the play of a computer program. In our scenario we still attempt to learn the weights of an evaluation function, but the objective is to imitate the style of a given player rather than improve the program's play. Of course, if the player under consideration is very strong, for example Kasparov or Kramnik, then the resulting program is likely to improve; but the method could also be used to learn the evaluation functions of weaker players.

\smallskip

The learning algorithm described in Section~\ref{sec:learn-eval}, based on Sutton's TD(0) \cite{SUTT88}, corresponds to the simplest rule, which updates only the current predictions.
We note that a more general formulation proposed by Sutton is TD($\lambda$); this utilises a decay factor $\lambda$ between 0 and 1, and forces the algorithm to also take into account earlier predictions. To accelerate the training, we utilise both an adaptive learning rate and a momentum term \cite{ALME97,REED99}, as we describe in Subsection~\ref{subsec:adapt}. In Section~\ref{sec:exp} we present a proof of concept, where we attempt to learn the styles of two recent Chess world champions, Kasparov and Kramnik, and we make use of the learnt feature weights to guess, in a game played between the two players, who was white and who was black. Despite some encouraging results, there are also some fundamental limitations of our approach for defining a player's ``style''. In particular, as pointed out to us by Chess Grandmaster Pablo San Segundo \cite{SANS08}, our choice of features (described in Subsection~\ref{subsec:setup}) is probably too low-level, since all strong players seek to optimise the  placement of their pieces and maintain a combination of pieces according to sound tactical and positional criteria. On a higher level, it is tempting to classify Kasparov as a more ``tactical'' player and Kramnik as a more ``positional'' player. However, these concepts are difficult to formulate in a precise manner and, moreover, it is not clear how to translate them into an algorithmic framework. We discuss these and other issues in Subsection~\ref{subsec:limits}.
In Section~\ref{sec:conc} we give our concluding remarks.

\section{Temporal Difference Learning of an Evaluation Function}\label{sec:learn-eval}

Temporal difference learning \cite{SUTT88} has been widely used to tune the evaluation function component of computer game playing programs \cite{LEVY91}, for example, in \cite{SAMU59,TESA92,SCHR94,LEOU95,BAXT00,BEAL00,SCHA01}. The evaluation function is the component of a computer game playing program that maintains the board features that are statically evaluated by the program. By combining state-of-the-art minimax tree search \cite{MARS00} and game specific heuristics, computer game playing programs have achieved world-class level, surpassing human performance in Backgammon, Othello, Checkers and Chess. It is noteworthy that computer Go programs still only play at amateur level, but employing recent advances in Monte Carlo methods appears to be a promising approach for improving their performance \cite{GELL08}.

\smallskip


From now on we will concentrate on Chess and we assume that the essence of a player's style can be described by the relative weighting of the features of an evaluation function. We will focus on the task of tuning the weights of the evaluation function using a collection of the player's game records. In the context of Chess, many useful features have been proposed \cite{HART89,BERL90,CAMP02}; however, as we will discuss in Subsection~\ref{subsec:setup}, the choice of features is not easy, and we have incorporated some novel features relating to pawn structures and influence areas within the board, in addition to the conventional ones. Tuning an evaluation function from game records in the context of improving a computer's performance is a well-know approach \cite{KOJI01,TSUR02,MANN04,STER06,SILV07}, but employing it in the context of learning a player's style is novel.
In Subsection~\ref{subsec:adapt} we show how to accelerate the training by adapting the learning rate and adding a momentum term.

\medskip

Let us assume, without loss of generality, that we are learning white's evaluation function, and
that an evaluation function $V$ defines the value of a game position $s$ as the
weighted sum of the values $v_i(s)$ for each game feature $i$, with weights $w_i$, i.e.
\begin{displaymath}\label{eq:eval}
V(s) = \sum_{i=1}^n w_i v_i(s),
\end{displaymath}
where the values $v_i(s)$ are measured in units of a hundredth of a pawn (i.e. the value of a single pawn is $100$). All weights are constrained to be positive, and the weight of the material balance feature is kept constant at $1$, so that all other weights are relative to material balance.
We use the term {\em feature vector} for the vector of weights $w_i$.

\smallskip

The problem of tuning the evaluation function is that of learning the values of the weights $w_i$
that maximise the number of correct predictions of moves made by the given player. Usually,
the objective is to tune the evaluation function of a game playing program in order to improve the
``strength'' of the program. The relative ``strength'' of a program can be measured by its performance when  playing against another program (which is often the previous version of the same program prior to
tuning its evaluation function).

\smallskip

We convert the value $V = V(s)$ for a game position $s$ into a win probability $P(V)$ by
applying the logistic function (also known as the sigmoid function) to $V$, i.e. \begin{equation}\label{eq:logistic}
P(V) = \frac{1}{1 + \exp(-\kappa V)},
\end{equation}
where $\kappa$ is a constant, chosen here to be $0.01$.

\smallskip

The learning rule we use for adjusting the weights $w_i$ is the {\em delta learning rule} for perceptrons
\cite{REED99}. We assume that initially $w_i = 1$ for all $i$, i.e. all the features are assigned equal weights.

\smallskip

Let $s$ be a game position with white to move, and let $y = P(V(s))$ be white's win probability. (Recall that we have assumed we are learning the evaluation function from white's perspective.)
Now let $s'$ be the position with white to move after white's and black's next moves have been replayed from the game record, and let $z = P(V(s'))$ be the win probability for $s'$. (In other words, $s'$ is the resulting position after two further ply have been replayed from the game record.) The weights at time $t$ are updated using gradient descent, according to the following formula:
\begin{equation}\label{eq:delta}
\Delta w_i(t) = \frac{\eta (z - y)}{\kappa} \frac{\partial y}{\partial w_i},
\end{equation}
where $\eta > 0$ is a small positive constant, called the {\em learning rate}, and
\begin{displaymath}
\frac{\partial y}{\partial w_i} = \kappa y (1-y) v_i.
\end{displaymath}
(We note that $\kappa y (1-y)$ is the derivative of the logistic function.)

\smallskip

In \cite{MANN04}, the learning rate $\eta$ was set to $0.001$, although a learning rate of $0.1$ is often
recommended in the literature \cite{REED99}. In our experiments, we chose $\eta = 0.01$ as the initial learning rate (see Subsection~\ref{subsec:adapt} for more details).

\medskip

After each time they are updated, the weights could be normalised so that they sum to $1$, but we preferred simply to fix the weight of the material balance feature at $1$. The logic underlying this
decision is that it is customary to measure the value of a Chess position in terms of pawn units.
So, for example, a positional advantage can outweigh a deficit in material. In ``quiet''
positions, where there are no hidden tactics and the positional factors are balanced, the value of a
position can be measured by the material balance of the pieces on the board. In practice, the
material balance of a position often dominates the evaluation function $-$ but Chess would not be an
interesting game if this were always the case.

\smallskip

We note that the rule (\ref{eq:delta}) is a TD(0) temporal difference update rule
\cite{SUTT88}, since $(z - y)$ is the difference between the win probability $z$ after the player's and opponent's moves have been made and the win probability $y$ of the position before the moves are made.
There are two possibilities when evaluating the win probability $y$: (i) the minimax move that the program would choose is the same as the actual move made by the player from the game record, or (ii) the program would choose a different move. In case (i) the adjustments made to the weights are the same as they would be in self-play, the assumption being that predictions become more accurate as the game progresses. In case (ii) the adjustments made to the weights are such that the program will tend to more closely reflect the style of the actual moves made by the player.

\subsection{Adapting the Learning Rate and Adding a Momentum Term}\label{subsec:adapt}

A typical value used for the learning rate is $\eta = 0.01$, but we can also consider
individual adaptive learning rates $\eta_i(t)$ for each weight $w_i(t)$. The method
we used is similar to that in \cite{ALME97} (see also \cite{REED99} for related methods), which uses multiplicative increases and decreases of the rates. These are specified by
\begin{displaymath}\label{eq:lr}
\eta_i(t) = \left\{
\begin{array}{r@{\quad \quad}l}
u \eta_i(t-1) & {\rm if} \ \Delta w_i(t) \Delta w_i(t-1) > 0 \\
d \eta_i(t-1) & {\rm if} \ \Delta w_i(t) \Delta w_i(t-1) < 0 \\
\eta_i(t-1) & {\rm otherwise \ (no \ change),}
\end{array}
\right.
\end{displaymath}
where the constants $u > 1$ and $0 < d < 1$ control the rate of increase and
decrease, respectively; typically one takes $u \approx 1.1$ and $d \approx 0.9$. We restricted the learning rates so that the minimum allowed value was $0.01$ and the maximum was $1$; initially they were set at the minimum $0.01$.

\smallskip

We smoothed the gradient by adding a momentum term \cite{ALME97} (see also \cite{REED99}),
by setting
\begin{displaymath}\label{eq:momentum}
\phi_i (t) = \frac{(z - y)}{\kappa} \frac{\partial y}{\partial w_i} + \alpha \phi_i (t-1),
\end{displaymath}
where $0 \le \alpha < 1$ is the momentum parameter. Typically $\alpha$ is between $0.5$ to $0.95$
\cite{ALME97}; we chose $\alpha$ to be $0.6$. The update rule (\ref{eq:delta}) is now modified to
\begin{displaymath}
\Delta w_i(t) = \eta_i  (t) \phi_i (t).
\end{displaymath}

We note that the momentum can also be viewed as giving the procedure memory that decays over time, somewhat akin to the more general TD($\lambda$).

\section{Proof-of-Concept Experiment}\label{sec:exp}

In the following subsections we describe a proof-of-concept experiment, where our task was to learn the styles of two recent Chess world champions, Kasparov and Kramnik. The resulting evaluation functions were tested by trying to discriminate between the two players from records of games between them.

\smallskip

In Subsection~\ref{subsec:setup} we describe the components of the underlying Chess program used in the experiment, and in Subsection~\ref{subsec:eval} we describe the evaluation methodology we used to determine how well the learned evaluation functions discriminate between the two players. In Subsection~\ref{subsec:results} we discuss the results, and in Subsection~\ref{subsec:limits} we consider the limitations of our experiment and suggest how further progress can be made.

\subsection{Experimental Setup}\label{subsec:setup}

In order to carry out the experiment to learn the feature weights for a player's evaluation function, we first implemented a Chess playing program in Matlab. A comparable implementation in a programming language like C (possibly using open-source software) would be considerably faster (and thus allow deeper searches); however, we chose to use Matlab, firstly because of its convenience for experimentation in the early stages of working on the problem, but also for the challenge of implementing a Chess program in Matlab. The computations were carried out using Windows XP, running on a desktop PC with a 3.6 GHz Intel Pentium 4 processor and 2 GB of RAM.

\medskip

The components of the program included:
\renewcommand{\labelenumi}{(\alph{enumi})}
\begin{enumerate}
\item A parser for inputting moves from PGN (Portable Game Notation) files containing the game records.

\item A bitboard representation of the Chess board \cite{CRAC84}, and a bitboard move generator \cite{FENN08}.

\item A tree-search module, which implements the widely used NegaScout variation of the alpha-beta pruning minimax algorithm \cite{REIN83}.  The implementation includes quiescence search and a transposition table \cite{BREU97}.

\item An evaluation function that returns the value of a game position.
\end{enumerate}

For testing the learning algorithm, we chose 140 features:
\renewcommand{\labelenumi}{(\roman{enumi})}
\begin{enumerate}
\item The first 13 features were: material balance, pseudo-mobility \cite{HART89}, piece-square value tables from \cite{BEAL00}, having a bishop pair \cite{KAUF99}, having a knight pair, preference for a single bishop over a single knight, preference for a single knight over a single bishop, king safety in the form of having castled (with queens on the board), non-aversion to doubled pawns, preference for saddling the opponent with doubled pawns \cite{KAUF05}, having a queen-side majority, having a king-side majority, and the relative expansion factor. (The expansion factor is an idea of Chess Master Alexander Shashin, and is computed as the sum over the ranks of the number of the player's pieces on the rank multiplied by the rank. The relative expansion factor is the difference between the expansion factors for the two players \cite{SAVI04}.)

\item The next 9 features were related to 9 complexes of squares, defined by the four corners of the surrounding rectangle of each complex; for each complex, we measure the preference for (or aversion to)  occupying (or the opponent occupying) the complex. The complexes are: (1) a1,a3,c1,c3, (2) d1,d3,e1,e3, (3) f1,f3,h1,h3, (4) a4,a5,c4,c5, (5) d4,d5,e4,e5, (6) f4,f5,h4,h5, (7) a6,a8,c6,c8, (8) d6,d8,e6,e8 and (9) f6,f8,h6,h8.

\item The next 112 features relate to the preference for 112 adjacent pawn structures.

\item The final 6 features are: (1) isolated d-pawn, (2) no c-pawn and a non-isolated pawn on d4, (3) no e-pawn, a c-pawn and a non-isolated pawn on d4, (4) the Maroczy bind (pawns on c4 and e4 with no d-pawn), (5) no d-pawn and no pawn on c4 but a pawn on e4, and (6) semi-open c-file, i.e. no c-pawn but a non-isolated d-pawn not on d4.
\end{enumerate}

We note that our choice of features could be viewed as a limitation, since it is debateable whether they can adequately capture a player's style \cite{SANS08}.  This is discussed in Subsection~\ref{subsec:limits}.

\medskip

We close this subsection by mentioning a few practical considerations:

\begin{itemize}
\item For training purposes we considered only moves 5-35 from a game record in order to avoid early opening and endgames moves, which are normally dealt with using pre-computed lookup tables and separate evaluation functions.

\item For computational reasons the program performs a minimax search only to a depth of three ply, with check extensions and quiescence search taking into account all captures and checks at the first ply.

\item As we were concentrating on Kasparov's and Kramnik's evaluation functions, training was carried out using a collection of 1967 of Kasparov's games and 1738 of Kramnik's, and validation was carried out using 123 games between Kasparov and Kramnik.
\end{itemize}

\subsection{Evaluation Methodology}\label{subsec:eval}

The standard evaluation technique of using separate training and validation sets \cite{MITC97} was employed. We trained the weights of the evaluation functions for the two players, $S$ (Kasparov) and $M$ (Kramnik), using random selections of 1000 of each of their games. Testing was done using the entire validation set of 123 games.

\smallskip

We measure the absolute error between the current position $s$ and the position $s'$ resulting after another two ply from the game record $g$ have been replayed as
\begin{equation}\label{eq:abs}
e(s, g, p) = |P(V_p(s')) - P(V_p(s))|,
\end{equation}
where $P(V_p(\cdot))$ is the estimated win probability as defined in (\ref{eq:logistic}), and $V_p$ is the evaluation function trained for player $p$. This measure is natural in this context since it is precisely this quantity that temporal difference learning, as defined in (\ref{eq:delta}), is attempting to minimise.

\smallskip

Given a game $g$ and a player $p$, let $W(g)$ be the set of positions considered in the game $g$ where white is to move, and let $B(g)$ be those where black is to move. We assume, without loss of generality, that we are considering the game from white's perspective, whether $p$ is white or black.
The absolute error for the game is then defined as
\begin{equation}\label{eq:wg}
E(g, p) = \sum_{s \ in \ W(g)} e(s, g, p).
\end{equation}

\noindent We emphasise that $p$ refers to the evaluation function $V_p$ in (\ref{eq:abs}) and may or may not be the player that was actually playing white in $g$.

\smallskip

We define the {\em mean absolute error} (MAE) of the feature vector to be $E(g, p)$ divided by the number of positions in $W(g)$.


\smallskip

Assuming, without loss of generality, that $S$ was white in the game $g$, we classify $g$ as a {\em hit} for player $S$ with opponent $M$ if
\begin{equation}\label{eq:diff-mae}
E(g, M) - E(g, S) > \epsilon,
\end{equation}
where $\epsilon \ge 0$ is a threshold value, i.e. if we can correctly identify $S$ as white in the game $g$ because the absolute error for $g$ is less with $S$ playing white than with $M$ playing white.
If $S$ was actually black, the definition is still valid provided we consider the game from black's perspective, i.e. if $W(g)$ is replaced by $B(g)$ in (\ref{eq:wg}).

\smallskip

The {\em hit ratio} $H$ for player $S$ with opponent $M$ is defined as
\begin{displaymath}
H(S, M) = \frac{\# \{g \ in \ C \ | \ g \ {\rm is \ a} \ hit \ {\rm for} \ S \ {\rm with \ opponent} \ M \}}{\# C},
\end{displaymath}
where $C$ is the validation set of test games played between $S$ and $M$,
and $\# C$ is the cardinality of  $C$ (cf. \cite{MYSL97}); the hit ratio can be viewed as a measure of classification accuracy.

\smallskip

We emphasise that if $S$ was white in $g$ then the summation in (\ref{eq:wg}) is taken over $W(g)$, but if $S$ was black it is taken over $B(g)$; thus $H(S,M)$ is computed from $S$'s perspective, i.e. from white's perspective if $S$ was white and from black's perspective if $S$ was black.
In general $H(M, S) \not= H(S, M)$, since $H(M, S)$ is computed from $M$'s perspective.
We are therefore able to distinguish $S$'s and $M$'s styles if both $H(S, M) > 0.5$ and $H(M, S) > 0.5$ by a specified margin.

\smallskip

We note that if, for example, $H(S, M) > 0.5$ but $H(M, S) < 0.5$, then the classifier can discriminate between the players when the games are examined from $S$'s perspective, but not when they are examined from $M$'s perspective. This situation is obviously undesirable since, when attempting to classify a new game between the two players, we do not have the benefit of knowing in advance which player was white and which was black.

\subsection{Results}\label{subsec:results}

\begin{figure}
\centerline{\includegraphics[width=15cm,height=10cm]{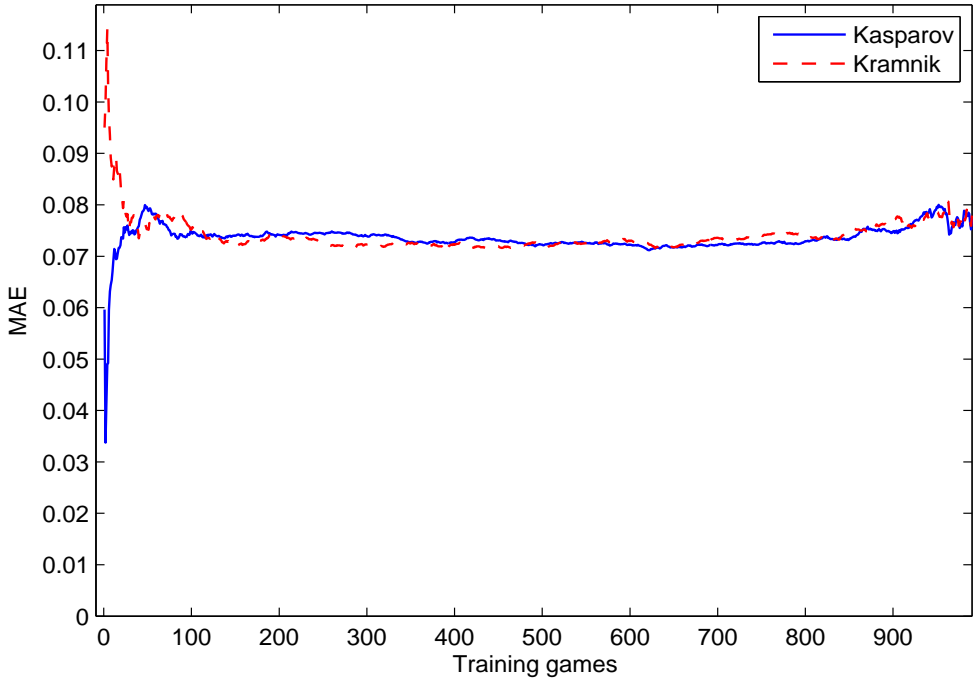}}
\caption{\label{fig:mae} Moving averages of the MAE of the feature vectors during training for Kasparov and Kramnik}
\end{figure}

We trained and tested our algorithm on the games of Kasparov and Kramnik, as described at the end of Subsection~\ref{subsec:setup}. Figure~\ref{fig:mae} shows the moving averages of the MAE of the feature vectors during training. We see that, after the first 50 or so games, the MAE is relatively stable and is quite similar for the two players. It is important that the MAEs do not differ by too much, in order to avoid any bias in the testing; in these tests the difference between the MAE of the two vectors over the training period was, on average, approximately $5 \times 10^{-5}$, i.e. less than $0.1\%$.

\smallskip

\begin{figure}
\centerline{\includegraphics[width=15cm,height=10cm]{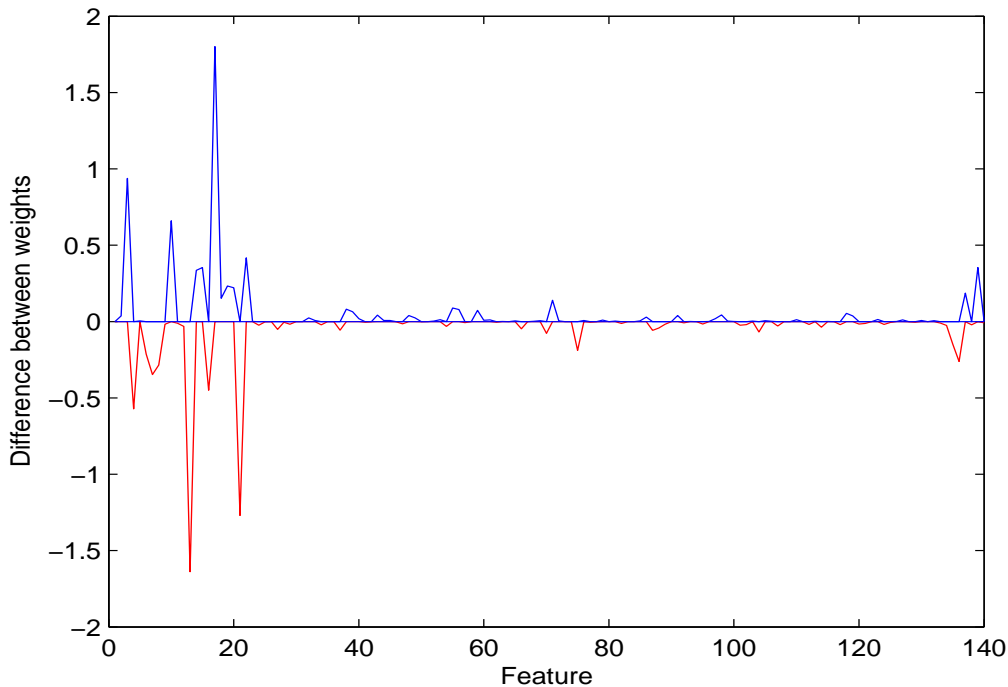}}
\caption{\label{fig:weights} The difference between Kasparov's and Kramnik's feature vectors}
\end{figure}

Figure~\ref{fig:weights} shows the difference between the feature vectors of the two players, where positive values indicate features for which the weights are higher for Kasparov's vector and negative values features for which they are higher for Kramnik's vector. There are four features for each player for which that player has the higher weight and the difference is greater than $0.4$; we now briefly discuss these. For Kapsparov, they are: the piece-square value tables, preference to saddle the opponent with doubled pawns, and the two complexes defined by squares a1,a3,c1,c3 (white's queen-side) and squares f6,f8,h6,h8 (black's king-side). The difference in weight for the piece-square value tables may be due to these being learnt from self-play \cite{BEAL00}, where games are generally decided as a result of tactical play, which is closer to Kasparov's highly dynamic style. The weight differences for the two complexes may indicate Kasparov's tendency as white to attack black's king, which normally castles on the king-side, and as black to opt for an attack on white's king when the players castle on opposite wings. For Kramnik, the four features are: preference for the bishop pair, the relative expansion factor, and the two complexes defined by squares f4,f5,h4,h5 (the central king-side) and squares f1,f3,h1,h3 (white's king-side). The relative expansion factor and the preference for the central king-side may be related to Kramnik's preference for manoeuvering on the king-side, and the preference for white's king-side may indicate his preference for keeping his king safe, especially when he is white. Regarding the pawn structure features, there is only one for each player for which the weight difference is greater than $0.25$. For Kasparov, it is feature (iv)(5) in Subsection~\ref{subsec:setup} (with a difference of $0.3531$), which may indicate his preference for a pawn on e4 and the absence of a pawn on d4. For Kramnik, it is feature (iv)(2) in Subsection~\ref{subsec:setup} (with a difference of $0.262$), which may indicate his preference for a pawn on d4 and the absence of a pawn on c4. These differences may reflect their preferred openings, since these often determine the middle game pawn structure.

\smallskip

Although these observations are interesting, it is clear that, as discussed above, the features we are using are not sufficient to fully capture the different styles of the two players.
Moreover, the weights on their own do not tell the full story, as some features may tend to have higher values than others. In our case, however, apart from material balance (which has a fixed value), the values of all the other features are normally less than the value of a pawn. Nevertheless, in this context, feature selection, i.e. determining the dominant features in each player's evaluation function, may be useful.

\medskip

In order to optimise the results, we chose to test the trained feature vectors just on moves 25 to 35 from the validation set of 123 games. This choice was motivated by the fact that we expected the differences in style to be most noticeable in proper middle game positions. In particular, we were not
attempting to capture their individual opening preferences, which are easily detected at the meta-level, for example, by comparing opening sequences to an opening book database. Nevertheless, the choice of opening does reflect style to some degree and pawn structures often persist until the endgame. As pointed out to us by Chess Grandmaster Pablo San Segundo \cite{SANS08}, the choice of opening does not always correlate with style as there may be other considerations when choosing an opening, such as playing against a specific opponent or the tournament situation of the player.

\smallskip

\begin{figure}
\centerline{\includegraphics[width=15cm,height=10cm]{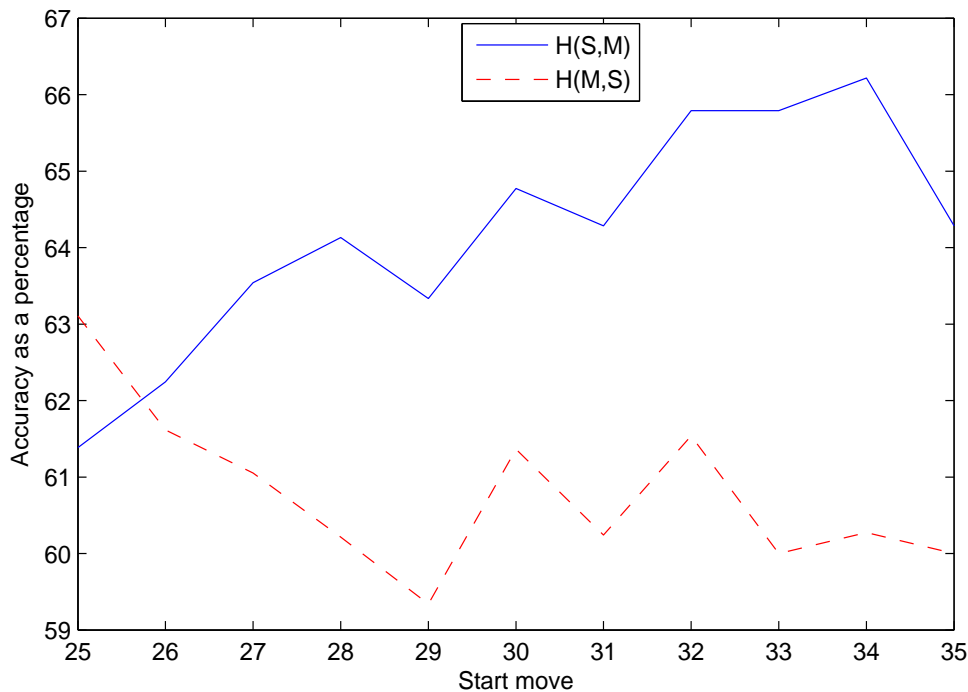}}
\caption{\label{fig:class} Classification accuracy for games between Kasparov and Kramnik}
\end{figure}

In Figure~\ref{fig:class} we show $H(S, M)$ as the continuous line and $H(M, S)$ as the broken line, where the threshold $\epsilon$ was set to zero, the start move was varied from 25 to 35, and the end move was fixed at 35. The mean of $H(S, M)$ is $0.642$, and the mean of $H(M, S)$ is $0.608$, which clearly shows the potential of the method. Moreover, we note that the mean value of the difference between the MAE for $S$ and $M$ from $S$'s perspective when counting the hits for $H(S, M)$ is $0.0123$, while from $M$'s perspective when counting the hits for $H(M, S)$ it is $0.0128$.

\medskip

Despite this moderate success, we could not replicate this result for the games of Topalov ($T$), another former world champion, under the same training regime. Although we obtained the value of  $0.645$ as the mean of $H(S, T)$, the very low value of  $0.283$ was obtained for $H(T, S)$. Correspondingly, although we obtained the value of $0.603$ as the mean of $H(M, T)$, the low value of $0.405$ was obtained for $H(T, M)$. It is possible that 1000 games are not enough to train the weights for Topalov's feature vector. Evidence for this is that the average differences during training between the MAE of Topalov's feature vector and both  Kasparov's and Kramnik's was approximately $6 \times 10^{-3}$, i.e. more than $7\%$. However,
as mentioned above, the average difference between the MAE of Kasparov's and Kramnik's feature vectors was only $5 \times 10^{-5}$, less than $0.1\%$. Moreover, the MAE of Topalov's feature vector was diverging rather than converging as the training increased. The failure to train an adequate feature vector for Topalov may partly be due to the limitations of the features we have selected, and possibly also to other limitations of our approach, as discussed in the next subsection.

\subsection{Limitations}\label{subsec:limits}

As noted in the introduction, our choice of features for classifying players' styles is probably too low-level, since strong players will normally play strong moves in any position \cite{SANS08}.
It is possible that a higher level abstraction of a player's style would emerge from a substantial increase in the number of features (Deep Blue had approximately 8000 features \cite{CAMP02}), given a sufficient increase in computing power. An example of such emergence is the ``positional'' 37th move (Be4) played by Deep Blue against Kasparov in their rematch in 1997; this move unsettled Kasparov for the rest of the match, which he subsequently lost.

\medskip

We are unsure what the best approach may be for capturing higher level elements of playing style, such as ``positional'' versus ``tactical'', within an algorithmic framework.
One possible way forward for recognising positions as tactical may be indicated by the observation that tactical ability requires strong calculation. We note that a wide range of Chess manuals promote improvement of tactical ability through puzzles (many of which are available in electronic form) that can readily be solved with the aid of a powerful computer Chess program.
On the other hand, fewer puzzles for improving ``positional'' ability exist, and their solution often involves a {\em plan} rather than an individual move; such a solution, in the form of a plan, is not readily obtainable with the aid of current computer Chess technology, which puts the emphasis on brute-force calculation rather than on any form of planning. Another possibility is to design and include higher level features that better capture playing style, but we leave this as a possible direction for future research.

\section{Concluding Remarks}\label{sec:conc}

The aim of this research was to use machine learning to capture the style of human Chess players and use this knowledge to discriminate between players by inspecting records of games played between them.
We have presented some preliminary results using a conventional Chess engine architecture combined with the method of temporal difference learning. This has yielded some success, as described in Subsection~\ref{subsec:results}. Although we believe that the methodology we have presented is sound and potentially viable, we have also uncovered some fundamental issues that need to be addressed if further progress is to be made. In particular, it would be desirable to capture higher level concepts, such as ``tactical'' versus ``positional'', and to be able to classify the choices players make during a game according to the degree to which that they match these concepts.

\smallskip

Since methods used in the domain of Chess frequently transfer to other strategic board games, it would be interesting to try our approach on games such as Checkers and Go.
We conclude with the speculative suggestion that there may be even wider domains of application to, for example, learning profiles of agents from records of sequences of their actions.

\end{document}